\documentclass[letterpaper, 10 pt, conference]{ieeeconf}  

\IEEEoverridecommandlockouts                              

\usepackage{graphics} 
\usepackage{amsmath} 
\usepackage{amssymb}  

\usepackage{booktabs}
\usepackage{multirow}
\usepackage{graphicx}
\usepackage{caption}
\usepackage{threeparttable}
\usepackage{booktabs}
\usepackage{multicol}
\usepackage{float}
\usepackage{balance}
\usepackage{cite}
\usepackage{color}

\usepackage{hyperref}

\title{\LARGE \bf
UMAD: University of Macau Anomaly Detection Benchmark Dataset
}

\author{
Dong Li$^{1}$, Lineng Chen$^{1}$$^{2}$, Cheng-Zhong Xu$^{1}$, and Hui Kong$^{1*}$ 
\thanks{This work was supported in part by the Science and Technology Development Fund of Macau SAR under Grants 0046/2021/AGJ/ and 0067/2023/AFJ.}
\thanks{*Corresponding author.}
\thanks{$^{1}$The State Key Laboratory of Internet of Things for Smart City (SKL-IOTSC), Faculty of Science and Technology, University of Macau, Macao, China. {\tt\small \{lidong8421bcd, linengchen\}@gmail.com, \{czxu, huikong\}@um.edu.mo}}
\thanks{$^{2}$The Key Laboratory of Education Blockchain and Intelligent Technology, Ministry of Education, Guangxi Normal University, Guilin, China.}}

\begin{document}
\makeatletter
\g@addto@macro\@maketitle{
\centering
\setcounter{figure}{0}
\begin{center}
\vspace{0.0in}
    \centering
    \includegraphics[scale=0.95]{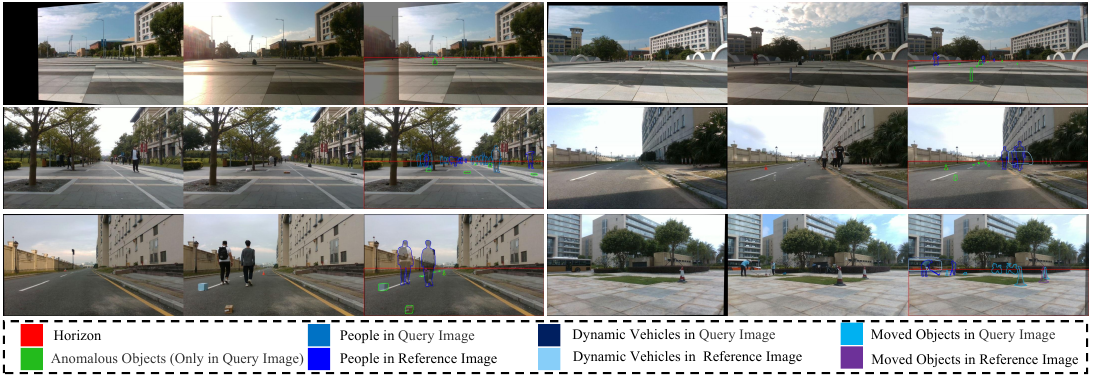}
\captionof{figure}{We introduce \textbf{\textit{UMAD}}, a large-scale reference-based anomaly detection dataset capturing real-world scenarios. \textit{UMAD} contains 6 distinct scenes, \textbf{120} sequences, \textbf{26k} image pairs, and a comprehensive set of \textbf{140k} object annotation labels. Featuring high-precision alignment and fine-grained annotation of images captured under diverse lighting conditions, \textit{UMAD} establishes a large comprehensive benchmark for the challenging task of reference-based anomaly detection. The third column images are the result of overlaying the first and second images, along with the contour of the mask.}\label{img8}
\vspace{-0.2in}
\end{center}
}
\makeatother
\maketitle
\setlength{\textfloatsep}{3pt}


\thispagestyle{empty}
\pagestyle{empty}

\begin{abstract}




Anomaly detection is critical in surveillance systems and patrol robots by identifying anomalous regions in images for early warning. Depending on whether reference data are utilized, anomaly detection can be categorized into anomaly detection with reference and anomaly detection without reference. Currently, anomaly detection without reference, which is closely related to out-of-distribution (OoD) object detection, struggles with learning anomalous patterns due to the difficulty of collecting sufficiently large and diverse anomaly datasets with the inherent rarity and novelty of anomalies. Alternatively, anomaly detection with reference employs the scheme of change detection to identify anomalies by comparing semantic changes between a reference image and a query one. However, there are very few ADr works due to the scarcity of public datasets in this domain.  In this paper, we aim to address this gap by introducing the UMAD Benchmark Dataset. To our best knowledge, this is the first benchmark dataset designed specifically for anomaly detection with reference in robotic patrolling scenarios, e.g., where an autonomous robot is employed to detect anomalous objects by comparing a reference and a query video sequences. The reference sequences can be taken by the robot along a specified route when there are no anomalous objects in the scene. The query sequences are captured online by the robot when it is patrolling in the same scene following the same route. Our benchmark dataset is elaborated such that each query image can find a corresponding reference based on accurate robot localization along the same route in the pre-built 3D map, with which the reference and query images can be geometrically aligned using adaptive warping. Besides the proposed benchmark dataset, we evaluate the baseline models of ADr on this dataset. We hope this benchmark dataset will facilitate the advancement of ADr methods in the future. Our UMAD benchmark dataset will be publicly accessible at \href{https://github.com/IMRL/UMAD}{https://github.com/IMRL/UMAD}.

\end{abstract}

\begin{table*}[thbp]
\centering
\caption{Comparison of the proposed dataset and existing representative scene change detection datasets}

\label{ex2}

\begin{tablenotes}
\item[a] \tiny * Not yet publicly released.
\item[b] \tiny $^{\ddagger}$ The UMAD dataset provides four semantic labels and indicates the images on which these labels are present.
\end{tablenotes}

\resizebox{\linewidth}{!}{
\begin{tabular}{c|cccccccccc}
\specialrule{1.5pt}{0pt}{0pt} 
\textbf{Datasets} & \textbf{Year} & \textbf{Real/Sim} & \textbf{Indoor/Outdoor} & \textbf{Sequences} & \textbf{Image Pairs} & \textbf{Resolution} & \textbf{Image Alignment} & \textbf{\#Semantic Class} & \textbf{Target change} & \textbf{Non-target change}\\
  \hline
  \hline
  CD2014 \cite{wang2014cdnet} & 2014 & Real & Outdoor & 1 & 70000 & 720 $\times$ 480 & - & 1 & dynamic object & -\\
  PCD TSUNAMI \cite{jst2015change} & 2015 & Real & Outdoor & 1 & 100 & 1024 $\times$ 224 & - & 1 & structural change & weather, light\\
  PCD GSV \cite{jst2015change} & 2015 & Real & Outdoor & 1 & 100 & 1024 $\times$ 224 & - & 1 & structural change & weather, light\\
  VL-CMU-CD \cite{alcantarilla2018street} & 2018 & Real & Outdoor & 152 & 1362 & 1024 $\times$ 768 & coarse & 4 & structural change & weather, light\\
  PSCD \cite{sakurada2020weakly} & 2020 & Real & Outdoor & 1 & 770 & 4096 $\times$ 1152 & - & 8 & structural change & weather, light\\
  CARLA-OBJCD* \cite{hamaguchi2020epipolar} & 2020 & Sim & Outdoor & 1 & 15000 & - & - & 10 & new/missing object & light\\
  GSV-OBJCD* \cite{hamaguchi2020epipolar} & 2020 & Real & Outdoor & 1 & 500 & - & - & 10 & new/missing object & weather, light\\
  Changesim \cite{park2021changesim} & 2021 & Sim & Indoor & 20 & $\sim$130,000 & 640 $\times$ 480 & - & 24 & new/missing/rotated/replaced object & dusty air, low-illumination\\
  Standardsim \cite{mata2022standardsim} & 2022 & Sim & Indoor & - & 12718 & 1280 $\times$ 720 & - & - & new/missing object & -\\
  \hline
  \midrule
  \multirow{2}{*}{\textbf{UMAD (Ours)}} & \multirow{2}{*}{\textbf{2024}} & \multirow{2}{*}{\textbf{Real}} & \multirow{2}{*}{\textbf{Outdoor}} & \multirow{2}{*}{\textbf{120}} & \multirow{2}{*}{\textbf{26301}} & \multirow{2}{*}{\textbf{1280 $\times$ 720}} & \multirow{2}{*}{\textbf{fine}} & \multirow{2}{*}{\textbf{4 (7)$^{\ddagger}$}} & \textbf{anomalous object,} & \multirow{2}{*}{\textbf{weather, light}} \\
  & & & & & & & & & \textbf{dynamic object, new/missing object} & \\
  \bottomrule
  
  \specialrule{1.5pt}{0pt}{0pt}
\end{tabular}
  }
 \vspace{-10pt}
\end{table*}

\section{Introduction}

Anomaly detection (AD) is a task focusing on identifying anomaly regions from given images, which is one of the important applications for surveillance systems and patrol robots. Generally, anomaly detection can be categorized into anomaly detection with reference (ADr) and anomaly detection without reference (ADwr) depending on whether using reference data or not\cite{jiao2023survey}.

ADwr is closely related to out-of-distribution (OoD) object detection, aiming to detect the anomalous parts that are out of the normal distribution~\cite{jiang2022survey}. However, this remains a challenging task as most state-of-the-art (SOTA) semantic segmentation methods, trained in a closed-set manner, often fail to detect unknown anomalies beyond the pre-defined classes. Furthermore, due to the inherent rarity and novelty of anomalous objects, it is difficult to collect sufficiently large and diverse anomaly datasets to facilitate a comprehensive learning of anomalous patterns. To increase the diversity of anomalies, several anomaly datasets such as Fishyscapes~\cite{blum2021fishyscapes} and CAOS~\cite{hendrycks2022scaling}, choose to augment their OoD samples by pasting items from public datasets (e.g. COCO or Pascal VOC) into regular scenes as anomalous objects. However, the synthetic data cannot fully represent what may occur in real-world scenarios. To alleviate the dependence on the diversity of the anomalies, some recent promising anomaly segmentation methods~\cite{lis2019detecting,di2021pixel} adopt generative models to re-synthesize photo-realistic images from the predicted semantic map and then locate anomalies by comparing the differences between the original image and the reconstructed one which can better preserve the appearance of known items compared to unknown anomalies. Alternatively, ~\cite{tian2023unsupervised} introduced pretrained Multimodal Large Language Models (MLLMs), such as the Contrastive Language-Image Pretraining (CLIP)~\cite{radford2021learning} trained on 400 million text-image pairs collected from the internet, to improve the performance of predicting road anomalies by leveraging rich open-set multimodal semantic information. However, both generative- and MLLM-based methods suffer from high computational costs, making real-time deployment on mobile robots challenging.


In contrast to ADwr, which struggles with the diversity of anomalies in real-world scenarios, ADr simply employs the change detection (CD) approach to identify anomalies. Specifically, given an anomaly-free reference image and the query one that may contain anomalous objects, ADr can locate candidate anomalies by detecting semantic changes between these two images. For instance, ~\cite{kong2010detecting} can effectively implement anomaly detection of various abandoned objects by identifying the differences between the reference and target videos. Similarly, ~\cite{xie2023spectral} can realize anomaly detection in the field of remote sensing by calculating the difference between two hyperspectral images. However, despite not needing to tackle the diversity of anomalies, there are still few related ADr works because of the lack of corresponding public datasets. Although the widely used scene change detection (SCD) datasets, such as VL-CMU-CD~\cite{alcantarilla2018street} and CDnet series~\cite{goyette2012changedetection,wang2014cdnet}, can be applicable, they mainly focus on evaluating models for detecting changes not anomalies in dynamic urban scenes. These changes typically involve the addition or removal of landmarks, pedestrians, vehicles, and other roadside buildings\cite{shi2020change}, not specifically labeled for various anomalies that can be used for anomaly detection. For an overview of the current scene change detection datasets, refer to Table \ref{ex2}.

In this paper, we propose an ADr benchmark, the University of Macau Anomaly Detection (UMAD), as shown in Fig. \ref{img8}. To the best of our knowledge, we are the first to collect the ADr dataset for the robotic patrolling scene, enabling the development and evaluation of ADr methods. 
Specifically, reference and query sequences in UMAD are captured by a mobile robot mounted with a camera along a predetermined route at different times. This setup allows for the evaluation of ADr models in handling diverse non-semantic changes that occur in real-life scenarios, including variations in illumination, the presence of shadows, and the impact of camera back-lighting. At the same time, the presence of dynamic objects in the patrol scene can assess the performance of the ADr models in locating anomalous changes rather than focusing solely on semantic changes.
To mitigate the impact of viewpoint changes between the image pairs, the corresponding reference and query images are selected based on robot localization in the prebuilt map of the scene. Then we propose a warping method to align the corresponding reference and query image pairs.

UMAD covers 6 different scenes in total, each consisting of over 9 raw sequences (including at least 3 reference sequences and 6 query sequences), resulting in over 26k labeled image pairs for training and evaluating the models. Note that reference images are anomaly-free and query images may or may not contain anomalies that simulate real patrol scenes. We manually annotate the pixel-wise ground truth of semantic changes below the horizon line in all image pairs, including the above-mentioned dynamic objects and artificially placed anomalous objects.
Given the absence of publicly available ADr models, our approach involves conducting ADr in a change detection manner. Notably, there are two approaches for anomaly detection using change detection. The first approach involves binary ADr, where anomalies are directly detected by comparing the differences between image pairs while disregarding non-anomalous changes, such as those caused by dynamic objects. The second approach is multi-class ADr, which utilizes change detection to locate changes and further distinguishes between anomalous and non-anomalous changes. By conducting experiments with these two types of anomaly detection methods, we aim to explore the suitability of these models for ADr and provide insights into the performance of these two approaches in accurately detecting anomalies, thus contributing to the development of an effective ADr algorithm.

In summary, the main contributions of our work are listed as follows:
\begin{itemize}
{
\item We propose a large comprehensive ADr dataset for robotic patrol applications under different lighting conditions, by which we hope to boost the research in anomaly detection with reference.

\item We propose an adaptive image warping method that approximately achieves pixel-wise alignment between the reference and query images to facilitate anomaly detection via change detection in the aligned reference and query images.

\item We conduct experiments based on the baseline models for anomaly detection on the UMAD dataset and reveal future feasible directions of anomaly detection research based on analyzing their performance.
}
\end{itemize}


In the rest of this paper, we first review the existing public SCD datasets and methods in Sec \ref{sec:2}. Next, we describe our UMAD benchmark in more detail in Sec \ref{sec:3}. Then we provide extensive experiments to evaluate the SOTA SCD methods on our benchmark in Sec \ref{sec:4}. Finally, we conclude our work in Sec \ref{sec:5}.

\section{Related Work}
\label{sec:2}

\subsection{Change Detection Algorithms}
Change detection aims to identify differences between image pairs.  Change detection algorithms are typically expected to exhibit robustness to temporal variations, such as changes in illumination and viewpoint, while also being capable of detecting semantic changes, such as the emergence or disappearance of objects. There are two main application scenarios based on change detection: remote sensing change detection (RSCD) and street-scene change detection (SCD) \cite{varghese2018changenet}. RSCD focuses on identifying changes on the Earth's surface based on remote sensing images, with applications in urban planning, environmental monitoring, disaster assessment, and so on \cite{shi2020change}. 

In contrast to RSCD, SCD methods are more relevant to our work and their scenarios are mostly street scenes. Traditional approaches in this field often employ handcrafted features and feature-matching techniques \cite{radke2005image}. With the advancements in deep learning, many change detection methods have adopted deep neural networks, leading to better performance. 
For instance, ~\cite{jst2015change} combined a convolutional neural network (CNN) for feature extraction to generate a low-resolution feature map, and utilize superpixel segmentation to get accurately change boundaries. ChangeNet~\cite{varghese2018changenet} adopted a siamese network~\cite{mueller2016siamese} and a fully convolutional network (FCN)~\cite{shelhamer2017fully} to extract features from image pairs and detect visual changes. CSCDNet~\cite{sakurada2020weakly} increased the depth of the network based on the ResNet block to improve model performance. CDNet++~\cite{prabhakar2020cdnet++} introduced five correlation layers at corresponding feature levels. HPCFNet~\cite{lei2020hierarchical} employed hierarchical feature combinations across multiple levels to handle multi-scaled objects and introduced a multi-part feature learning strategy to address the issue of unbalanced locations and scales of changed regions. DR-TANet~\cite{chen2021dr} proposed a temporal attention module to identify similarities within a fixed dependency scope and integrated concurrent horizontal and vertical attention mechanisms to refine strip entity changes. In contrast to previous works that struggled with general segmentation problems, C-3PO~\cite{wang2023reduce} proposed a novel paradigm that simplifies change detection by reducing it to semantic segmentation. This approach leverages the capabilities of powerful existing semantic segmentation networks to address general segmentation challenges in the change detection domain.

\subsection{Scene Change Detection Datasets}
CDnet2012~\cite{goyette2012changedetection} and CDnet2014~\cite{wang2014cdnet} are highly popular scene change detection benchmarks that offer complete pixel-wise ground truth. 
The CDnet series provides diverse scenarios, including corridors, parks, lakesides, bus stations, streets, highways, libraries, offices, and blizzards. These datasets are widely utilized for evaluating the generalization ability of change detection algorithms due to their data diversity and high-quality ground truth. 
Note that the cameras that collect images in these two datasets are usually static, so they do not need to align the reference and query image pairs. 


PCD dataset~\cite{jst2015change} consists of 200 pairs of panoramic images captured by the moving camera. These image pairs are selected from the closest viewpoints based on the GPS data to align the reference and query image pairs. 
So the PCD dataset does not align the reference and query image pairs well due to the large GPS uncertainty, which imposes extra requirements on subsequent change detection algorithms.

VL-CMU-CD dataset~\cite{alcantarilla2018street} is a street-view scene change detection dataset that spans a long period. 
This dataset mainly focuses on structural change for efficient map maintenance, so over 80\% of the change instances are bin, sign, vehicle, refuse, and construction in the street scene. 
VL-CMU-CD is the only existing real dataset with image alignment for scene change detection prior to our work. VL-CMU-CD designs a multi-sensor fusion Simultaneous Localization and Mapping (SLAM) system with a fast dense reconstruction system. However, this image alignment method disrupts the structure of one of the images, resulting in low alignment accuracy. Additionally, the computational cost of this alignment approach is high for online processing.

ChangeSim dataset~\cite{park2021changesim} is a photo-realistic dataset simulated for an indoor warehouse scenario. 
To establish correspondences between query and reference image pairs, the estimated pose of each query image is used, obtained through a visual SLAM algorithm RTABMAP~\cite{labbe2019rtab}. The changes in this dataset mainly involve the moving of industrial objects, such as pallets, forklifts, and pallet jacks. Although this dataset has pixel-level change labels, the simulated data are still quite different from the situation of the real scene. Recently, there are other simulated datasets such as CARLA-OBJCD\cite{hamaguchi2020epipolar} and Standardsim\cite{mata2022standardsim}. However, the environmental changes in these datasets are relatively simple.





\begin{figure}[thbp]
\centerline{\includegraphics[scale=0.5]{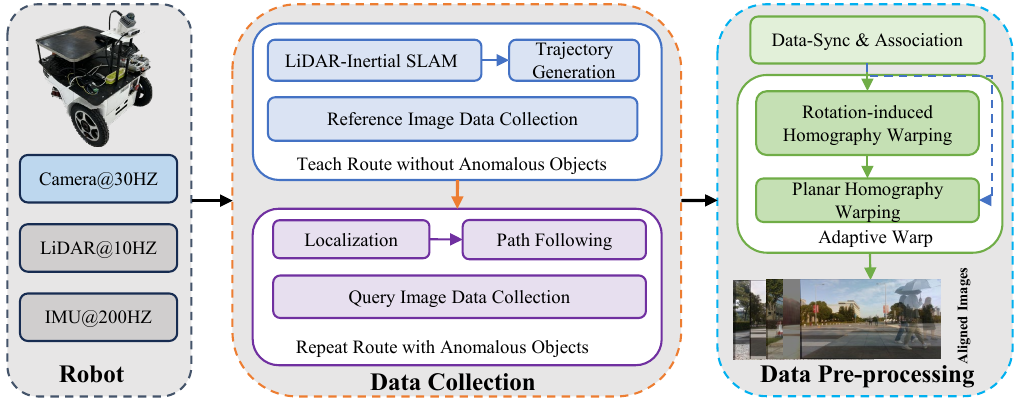}}
\caption{Overview of data collection and data pre-processing. }
\label{img1}
\end{figure}

\section{The UMAD Dataset}\label{sec:3}

The goal of our UMAD dataset is to introduce a new benchmark with high-precision geometric alignment between reference and query images to the community of Anomaly Detection. The dataset comprises six scenes collected by a mobile robot under various lighting conditions, consisting of 120 sequences, 26k image pairs and 140k labels categorized into seven semantic object classes.

Below we first introduce the robot platform for data collection when building UMAD, followed by an introduction to the data pre-processing, which mainly covers the adaptive warping method to achieve precise image alignment. Then we briefly introduced the process of data annotation. Finally, the statistics and properties of UMAD are elaborated. An illustration of the platform setup is shown in Fig. \ref{img1}. 

\subsection{Data Acquisition}

\noindent
\textbf{Sensor setup.} The UMAD dataset has been collected using the Giraffe ground robot as shown in Fig. \ref{img1}. The robot is controlled by a low-cost onboard computer with an Intel i7-1165G7 CPU. This platform has been equipped with the following sensors:

\begin{itemize}
\item (i) a Intel RealSense D435i camera capturing images with a resolution of 1280 $\times$ 720 pixels and at a frame rate of 30Hz.
\end{itemize} 

\begin{itemize}
\item (ii) a Livox Mid-360 solid-state LiDAR sensor with a 360° field of view (FOV), capturing point clouds at a rate of 10Hz, and a 200Hz IMU.
\end{itemize}

\begin{figure}[thbp]
\centerline{\includegraphics[scale=0.48]{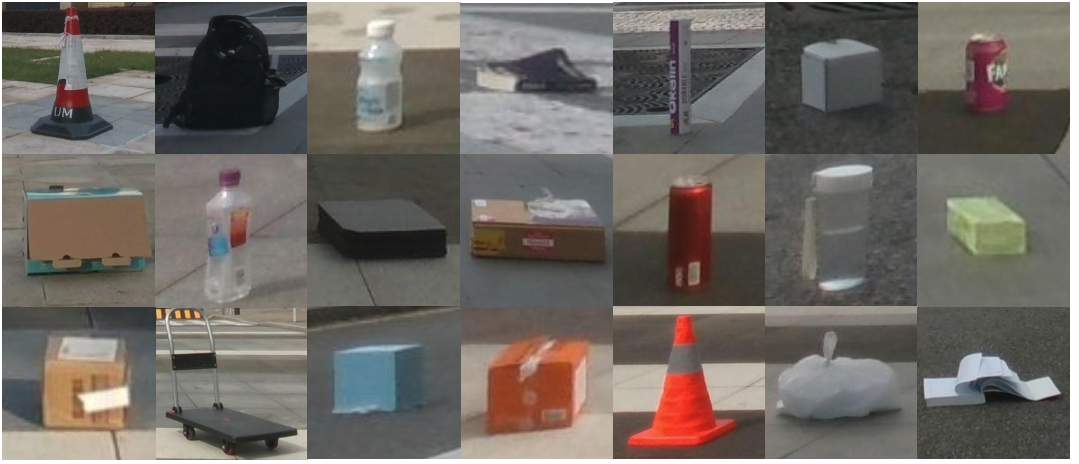}}
\caption{Examples of anomalous objects in the UMAD dataset.}
\label{img10}
\end{figure}

\vspace{-0.3cm}

\noindent
\textbf{Robot navigation system.} First, we manually control the robot to collect a teach-route and use the LiDAR-inertial SLAM \cite{xu2022fast} to estimate the robot's pose and build an accurate 3D point-cloud map of the scene. This sequence does not contain any anomalous objects and serves as the reference data. Next, utilizing the aforementioned 3D map and trajectory, 
we employ the pure pursuit  \cite{coulter1992implementation} to repeat the same route at a different time when the scene includes manually placed anomalous objects (as shown in Fig. \ref{img10}), and the presence of dynamic objects, such as people and vehicles, in the scene is typically considered normal. These newly collected data will be used as the query data.

\noindent
\textbf{Data collection.} To capture data from multiple sequences of the same scene, we collected reference and query data multiple times for each scene to enable a richer combination of data. Specifically, for each scene, we gathered at least three sets of reference sequences and six sets of query sequences. Consequently, each scene comprises a minimum of eighteen sequences. In total, the dataset is 14.5 km long with 1.52 km of unique paths.

\begin{figure}[thbp]
\centerline{\includegraphics[scale=0.47]{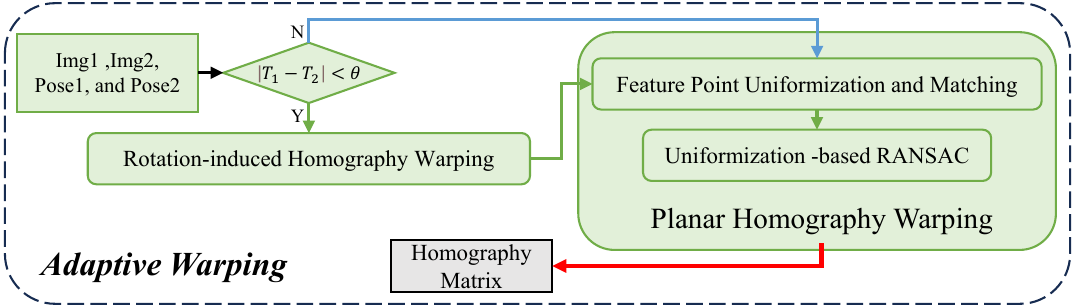}}
  \caption{Flowchart of Adaptive Warping. Img1 and Img2 represent a pair of reference and query images.}
\label{img3}
\end{figure}

\vspace{-0.3cm}

\subsection{Data Pre-processing}
%

\begin{figure*}[tbp]
\centerline{\includegraphics[scale=0.63]{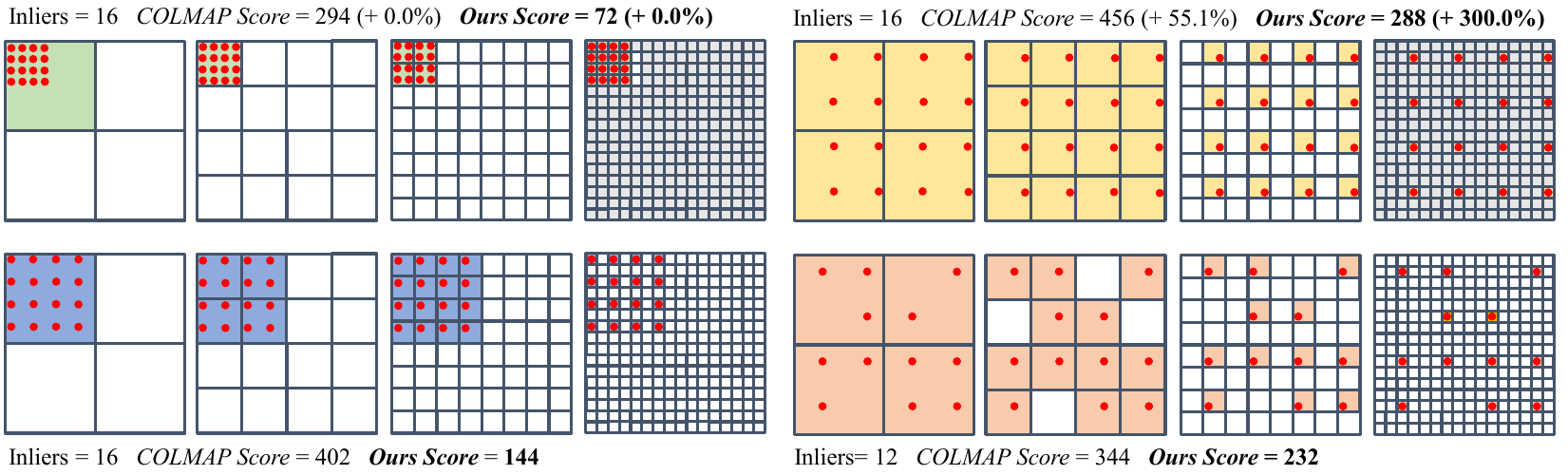}}
\caption{Comparison of our scores and COLMAP scores under different quantities and distributions of inliers in the image for L = 4.
In the first row, with 16 points, the ratio of our maximum score to the minimum score is four times. This indicates that our score exhibits a higher level of discrimination.
In the second row, compared to the scenario with 16 points concentrated on the left side, our score tends to favor a more evenly distributed scenario with 12 points on the right side.}
\label{img4}
\end{figure*}

With external calibration between the LiDAR and camera, we can readily get the camera's pose with respect to the world frame once we know the LiDAR sensor's pose, which can be obtained based on the LiDAR slam. Specifically, the LiDAR's pose of a reference sequence can be obtained in the teach stage based on the LiDAR full slam mapping procedure, whereas the LiDAR's pose of a query sequence can be obtained in the repeat stage based on the localization of the LiDAR sensor in the pre-built 3D map. Note that both the LiDAR and camera have been synchronized temporally.

For each query image, we can find its corresponding reference image based on their poses using the nearest-neighbor search.  
Next, we perform the adaptive warping between each corresponding pair of reference and query images to achieve high-precision image alignment, as illustrated in Fig. \ref{img3}. 

Let the poses of a pair of corresponding reference and query images $\{{I}_{1}, {I}_{2}\}$ be $\{{R}_{1}, {T}_{1}\}$ and $\{{R}_{2}, {T}_{2}\}$, representing the rotation matrix and translation vector, respectively. Let the corresponding camera intrinsic parameters be ${K}_{1}$ and ${K}_{2}$, which are given based on camera calibration.

\begin{figure}[thbp]
\centerline{\includegraphics[scale=0.53]{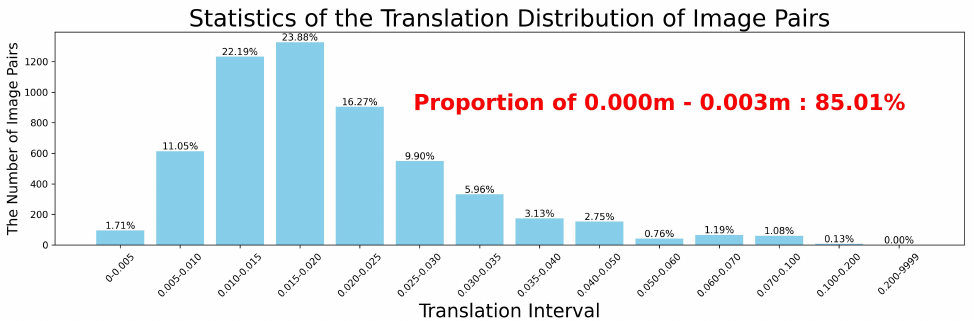}}
\caption{The translation statistics between the query images and reference images in a sequence of 5,556 raw data image pairs from Scene 1, Sequence 00.}
\label{img7}
\end{figure}

Fig. \ref{img7} illustrates that approximately 85\% of the pose translations between the corresponding reference and query image pairs are less than 0.03m, which means ${T}_{1}$ almost equals ${T}_{2}$. In addition, the majority of the corresponding orientations of ${I}_{1}$ and ${I}_{2}$ are also quite close to each other due to the high-precision localization in the pre-built map. But there are exceptions. In general, according to the difference between ${T}_{1}$ and ${T}_{2}$ and the difference between ${R}_{1}$ and ${R}_{2}$, we have two situations based on which we derive the adaptive warping scheme. 



\begin{itemize}
\item (i) A rotation-induced homography can be applied first to warp the two images if the relative translation is small enough (less than 0.03m for our case). Then apply a feature-based planar Homography warping to achieve accurate pixel-wise alignment. The reason for applying the planar homography is that the rotation-induced homography might not be perfectly estimated due to the pose estimation error. Therefore, the feature-based planar-homography warping is a good choice to remedy the rotation-induced homography warping. 
\end{itemize} 

\begin{itemize}
\item (ii) If the relative translations of the corresponding camera positions in the reference and query sequences are very different, it is better to directly apply a planar homography warping to the two images.
\end{itemize} 

We can justify the above adaptive warping scheme from the following perspective. In the planar homography transformation $m_{1} = \lambda {K}_{1}(R + \frac{1}{d} TN^{\top}) {K}_{2}^{-1} m_{2}$, where $\lambda$ is a scale factor, and $m_{1}=\left[u_{1}, v_{1}, 1\right]^{\top}$ and $m_{2}=\left[u_{2}, v_{2}, 1\right]^{\top}$ are the homogeneous coordinates of the corresponding pixels in the reference and query images.

$R={R}_{2} {R}_{1}^{\top}$ is the relative rotation between ${R}_{2}$ and ${R}_{1}$, and $T={T}_{2} -{T}_{1}$ is the relative translation between ${T}_{2}$ and ${T}_{1}$. The pixels $m_1$ and $m_2$ can be well aligned if the corresponding 3D point is in a planar region that fits the estimated homography, otherwise, the pixels $m_1$ and $m_2$ cannot be well aligned. However, when the relative translation $T$ is a small value (like in our case), the planar homography can be replaced by a rotation-induced homography, where the rotation is the relative rotation between ${R}_{2}$ and ${R}_{1}$. In that case, all pixels (not just the ones in the planar region) can be well aligned.

For us, the rotation-induced homography warping is based on the equation $m_{1} = \lambda {K}_{1}{R}_{2} {R}_{1}^{\top} {K}_{2}^{-1} m_{2}$ \cite{hartley2003multiple, ding2020homography}, where we do not need to apply the feature-based homography estimation process as in ORBSLAM \cite{mur2017orb} because we have already the ${R}_{1}$ and ${R}_{2}$ ready during localization. On the contrary, for the 
planar-homography estimation, we apply the feature-based scheme, where the distribution of feature corners (e.g., SIFT/SURF or ORB corners) can significantly impact the homography estimation's accuracy in general \cite{geiger2011stereoscan, mur2017orb}. To reduce the effect of the distribution of corner points, we first extract ORB feature points, and then perform uniformization and match them between the two images \cite{mur2017orb}.

Then in the RANSAC\cite{fischler1981random} stage, we aim to obtain a result with uniformly distributed inliers. Similar to COLMAP\cite{schonberger2016structure}, we discretize the image into a fixed-size grid with ${2}^{l}$ bins in both dimensions. The number of grids occupied in each layer is $k_l$, where $l = 1...L$ levels. In contrast, our approach accumulates the score over all levels with a weight that is dependent on the resolution, denoted as ${w}_{l} = {2}^{L- l + 1}$. Our score calculation is $Score = \sum_{l=1}^{L} {k_l \times 2^{L - l + 1}}$.
Fig. \ref{img4} illustrates the scores for different configurations and highlights the differences between our score and the COLMAP score.






\begin{table*}[tp]
  \centering
  \caption{The statistics of UMAD Dataset. Data for "Moved Objects," "People," and "Dynamic Vehicles" consists of a pair of reference and query images. }
  \label{tb1}
  \resizebox{\linewidth}{!}{
    \begin{tabular}{l|c|c|ccccc}
      \specialrule{1.5pt}{0pt}{0pt} 
      
      \textbf{Scenes \textbackslash Data} & \textbf{Sequences} & \textbf{Image Pairs} & \multicolumn{5}{c}{The number of labels (The average number of labels per image pair)} \\ 
      \cline{4-8}
       &  &  & \textbf{\#Anomalous Objects} & \textbf{\#People} & \textbf{\#Dynamic Vehicles} & \textbf{\#Moved Objects} & \textbf{\#All Objects}\\ 
      
      \hline
      1.N6 & 18 & 3422 & 12945 (3.78) & 5357 (1.57) & 411 (0.12) & 278 (0.08) & 18991 (5.55) \\
      2.Bridge & 24 & 5112 & 13687 (2.68) & 3693 (0.72) & 821 (0.16) & 340 (0.07) & 18541 (3.63)\\
      3.Central-Avenue & 18 & 5018 & 15268 (3.04) & 27704 (5.52) & 0 (0.00) & 2383 (0.47) & 45355 (9.04) \\
      4.Border-Road-1 & 18 & 3112 & 15598 (5.01) & 2321 (0.75) & 1952 (0.63) & 588 (0.19) & 20459 (6.57) \\
      5.Border-Road-2 & 18 & 4184 & 17333 (4.14) & 1233 (0.29) & 1701 (0.41) & 2380 (0.57) & 22637 (5.41)\\
      6.N2 & 24 & 5453 & 110837 (1.99) & 3226 (0.59) & 2064 (0.38) & 2322 (0.43) & 18449 (3.38)\\
      \hline
      \textbf{In Total}& \textbf{120} & \textbf{26301} & \textbf{85668 (3.26)} & \textbf{43524 (1.65)} & \textbf{6949 (0.26)} & \textbf{8291 (0.32)} & \textbf{144432 (5.49)}\\
      
      \specialrule{1.5pt}{0pt}{0pt} 
    \end{tabular}
  }
\end{table*}

\begin{table}[tp]
  \centering
  \caption{The point matching errors (PME) of our method and all comparison methods on our UMAD-homo-eva dataset. "Ours(rot)" represents the use of rotation-induced homography warping only.}
  \label{table:PME}
  \resizebox{\linewidth}{!}{
    \begin{tabular}{ccccccccc}
      \specialrule{1.5pt}{0pt}{0pt} 
      
      1) & & Scen. 1 & Scen. 2 & Scen. 3 & Scen. 4 & Scen. 5 & Scen. 6 & Avg\\
      \hline
      2) & $\mathcal{I}_{3\times3}$ & 14.29 & 18.17 & 9.09 & 8.54 & 16.38 & 9.23 & 12.53\\
      \hline
      3) & ORB + RANSAC & 5.27 & 18.93 & 4.44 & 5.99 & 3.43 & 6.43 & 7.41\\
      4) & ORB + MAGSAC & 4.15 & 11.45 & 4.07 & 5.02 & 3.02 & 5.39 & 5.52\\
      5) & SIFT + RANSAC & 2.90  & 4.61 & 3.70 & 2.79 & 3.08 & 4.06 & 3.52 \\
      6) & SIFT + MAGSAC & 2.96 & 4.48  & 3.52  & 3.29  & 8.50 & 4.08 & 4.47 \\
      \hline
      7) & Ours(rot) & 4.46 & 6.30 & 4.87 & 4.92 & 4.75 & 4.92 & 5.03\\
      8) & \textbf{0urs} & \textbf{2.57}  & \textbf{3.54} & \textbf{2.79} & \textbf{2.60}  & \textbf{2.65}  & \textbf{3.92} & \textbf{3.01} \\
      \specialrule{1.5pt}{0pt}{0pt} 
    \end{tabular}
  }
\end{table}

\subsection{Data Annotation}

We employ the Segment Anything Model (SAM) \cite{kirillov2023segment} for assisting annotation, which involves automatically generating object masks using SAM in most cases and manually assigning labels. For cases where SAM's automatic segmentation is not accurate enough, we resort to manual annotation.

To focus solely on changes in ground areas, we have annotated the "Horizon" in the query images. The region below the horizon is considered the region of interest. Furthermore, the query images have been labeled with categories such as "Anomalous Objects", "People", "Dynamic Vehicles", and "Moved Objects".

We have only labeled "People," "Dynamic Vehicles," and "Moved Objects" in the reference images since the reference images are aligned with the query images, and we believe there are no anomalous objects in the reference images.

"Moved Objects" refers to the variations in the scene, rather than the anomalous objects that have been intentionally placed, such as appearing/disappearing trash bins, etc.

\subsection{Dataset Statistics}

UMAD aims to provide comprehensive and diverse coverage of anomaly detection benchmark datasets.  Table \ref{tb1} provides an overview of the UMAD Dataset. 
Particularly, scene "2. Bridge" exhibits prominent undulations on its road surface, scene "4.Border-Road-1" and scene "5.Border-Road-2" share a highly similar visual style. The average number of people per pair of images in scene "3. Central-Avenue" is the highest in the entire dataset, reaching \textbf{5.52}.

\section{Experiment}

\subsection{Images Aligement}\label{exp1}

\begin{figure*}[thbp]
\centering
\includegraphics[scale=0.95]{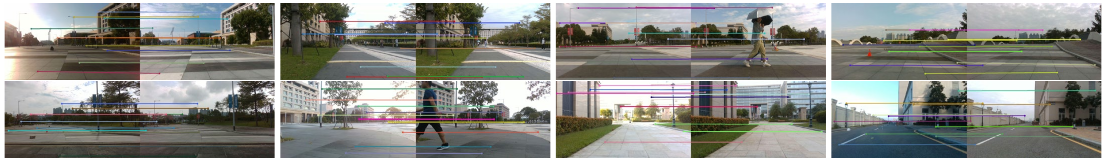}  
\caption{A glace of our \textit{\textbf{UMAD-homo-eva}} dataset. The figure showcases some illustrative examples of point correspondences for quantitative evaluation.}
\label{img6}
\vspace{-10pt}
\end{figure*}
\subsubsection{Dataset}
We present \textit{\textbf{UMAD-homo-eva}}, a small-scale evaluation dataset for homography estimation under various lighting variations, taking into account the absence of a dedicated dataset for this task. Our dataset consists of a total of 414 pairs of images across 6 different scenes. For each evaluated image pair, we manually annotated 10 uniformly distributed matching points for quantitative comparisons by pre-labeling them using traditional feature point extraction and matching methods. Some examples of our dataset are illustrated in Fig. \ref{img6}.

Following\cite{zhang2020content}, we use the point matching error (PME) as the evaluation metric. It adopts the average L2 distance between warped source points and target points for each pair of test images as an error metric. The calculation formula of PME is as follows: $PME = \frac{1}{N} \sum_{i=1}^{N} \|P_i - {P_i}^{{\prime}}\|_2$, where ${P_i}^{\prime}$ is the feature point warped by the estimated homography transformation. $P_i$ is the target ground-truth feature point. $N$ is the number of ground-truth feature point pairs.


\begin{table}[tbp]
  \caption{The label values for objects with different attributes in different tasks. Static means the background of the scene. Dynamic refers to common dynamic objects (common changes), such as people and cars. Anomaly refers to anomalous objects (anomalous changes), such as the anomalies introduced in Sec \ref{sec:3}.}
  \label{tab:4.1}
  \centering
  \vspace{-10pt}
    
    \vspace{5pt}
  \begin{tabular}{@{}lcccc@{}}
    \specialrule{1.0pt}{0pt}{0pt} 
    Tasks & Classes & Static & Dynamic & Anomaly \\ 
    \midrule
    Binary ADr  & 2 & 0 & 0 & 1 \\
    Multi-class ADr  & 3 & 0 & 2 & 1 \\
  \specialrule{1.0pt}{0pt}{0pt} 
  \end{tabular}
 
\end{table}

\subsubsection{Comparison of alignment performance}
We compare our method with some traditional feature-based solutions. Specially, we choose SIFT\cite{lowe2004distinctive}, ORB\cite{rublee2011orb} as the feature descriptors and choose RANSAC\cite{fischler1981random} and MAGSAC\cite{barath2019magsac} as the outlier rejection algorithms. We extracted 1000 points and set the RANSAC threshold to 4.0.
We compare our method with the others on our UMAD-homo-eva dataset, showcasing the quantitative results in Table \ref{table:PME}. 
\footnote{For more visualization results, please refer to the supplementary video}.

Using rotation-induced homography warping alone is superior to ORB but inferior to SIFT, and our ORB-based adaptive Warping approach performs better than both ORB and SIFT. Overall, our method reduces errors by 14.5\% compared to the second-best SIFT+RANSAC.


\begin{table*}[thbp]
  \caption{Results of different scene change detection methods on the UMAD dataset.
  }
  \vspace{-5pt}
  \label{tab:4.2}
  \centering
    \begin{tablenotes}
    \item[a]
    The $\uparrow$ and $\downarrow$ mean whether the anomaly detection results of the multi-class ADr are improved or decreased compared to the binary ADr, respectively.
    \end{tablenotes}
    \vspace{5pt}
  \begin{tabular}{@{}l|c|cccc|cccccccc@{}}
    \specialrule{1.0pt}{0pt}{0pt} 
    \hline
    \multirow{4}{*}{Methods} & \multirow{4}{*}{Backbone} & \multicolumn{4}{c|}{Binary} & \multicolumn{8}{c}{Multi-class}\\  \cline{3-14}
    

     ~ & ~ & \multicolumn{2}{c}{scene 2} & \multicolumn{2}{c|}{scene 3} & \multicolumn{4}{c}{scene 2} & \multicolumn{4}{c}{scene 3} \\ \cline{3-14}
    ~ & ~ & \multicolumn{4}{c|}{Anomaly} & \multicolumn{2}{c}{Dynamic} & \multicolumn{2}{c}{Anomaly} & \multicolumn{2}{c}{Dynamic} & \multicolumn{2}{c}{Anomaly}\\ \cline{3-14}
    
     ~ & ~ & IoU & F1  & IoU & F1 & IoU & F1 & IoU & F1 & IoU & F1 & IoU & F1 \\
    \hline
    FC-Siam-EF~\cite{{daudt2018fully}}  & U-Net & 10.2 & 18.6 & 8.7 & 16.1 & 11.3 & 20.1 & 21.1$\uparrow$ & 34.8$\uparrow$ & 41.1 & 58.3 & 21.7$\uparrow$ & 35.7$\uparrow$\\
    FC-Siam-diff~\cite{daudt2018fully}  & U-Net & 20.6 & 34.2 & 13.1 & 23.2 & 24.4 & 39.2 & 34.2$\uparrow$ & 50.9$\uparrow$ & 51.1 & 67.6 & 31.3$\uparrow$ & 47.6$\uparrow$\\
    FC-Siam-cov~\cite{daudt2018fully}  & U-Net & 20.7 & 34.3 & 14.3 & 25.1 & 16.2 & 27.9 & 28.0$\uparrow$ & 43.7$\uparrow$ & 41.6 & 58.7 & 26.7$\uparrow$ & 42.2$\uparrow$ \\    
    ChangeNet~\cite{varghese2018changenet}  & ResNet-50 & 38.2 & 55.3 & 37.6 & 54.7 & 22.8 & 37.1 & 37.3$\downarrow$ & 54.3$\downarrow$ & 34.0 & 50.7 & 35.5$\downarrow$ & 52.4$\downarrow$\\
    DR-TANet~\cite{chen2021dr} & ResNet-18 & 62.6 & 77.0 & 55.8 & 71.6 & 44.2 & 61.3 & 62.9$\uparrow$ & 77.2$\uparrow$ & 63.1 & 77.4 & 57.8$\uparrow$ & 73.3$\uparrow$\\
    CSCDNet~\cite{sakurada2020weakly}  & ResNet-18 & \textbf{65.1} & \textbf{78.9} & \textbf{59.6} & \textbf{74.7} & 58.6 & 73.9 & \textbf{67.2$\uparrow$} & \textbf{80.4$\uparrow$} & 73.1 & \textbf{84.4} & \textbf{63.5$\uparrow$} & \textbf{77.7$\uparrow$}\\
    C-3PO~\cite{wang2023reduce} & VGG-16 & 63.1 & 77.4 & 57.8 & 73.3 & \textbf{59.0} & \textbf{74.2} & 66.4$\uparrow$ & 79.8$\uparrow$ & \textbf{74.5} & 85.4 & 62.5$\uparrow$ & 76.9$\uparrow$\\
  \hline
  \specialrule{1.0pt}{0pt}{0pt} 
  \end{tabular}
  \vspace{-10pt}
\end{table*}
\vspace{-0.1cm}

\subsection{Anomaly detection with reference}
\label{sec:4}
In this section, to verify the effectiveness of UMAD, we adopt multiple open-source scene change detection models introduced in Section II to detect anomalies in the change detection manner. As listed in Table \ref{tab:4.1}, we conduct experiments related to two tasks: binary anomaly detection with reference, and multi-class anomaly detection with reference. We aim to find a more appropriate way to detect anomalies from the aligned pairs based on patrol robotics.

\subsubsection{Dataset split}
To establish the benchmarks, we first exclude the aligned pairs in which the query image does not contain anomalies and then split the rest into training, validation, and test sets according to the scenes. We selected 3 scenes ("1.N6", "5.Border-Road-2", "6.N2") as the training set (about 10.1k pairs), one scene ("4.Border-Road-1") for validation (about 2.7k pairs), and the remaining 2 scenes ("2.Bridge" and "3.Central-Avenue") as the test set (about 8.5k pairs).

\subsubsection{Evaluation metrics}
Following previous works~\cite{daudt2018fully,varghese2018changenet,sakurada2020weakly,chen2021dr,wang2023reduce}, we adopt Intersection over Union (IoU) and F1-score metrics to evaluate the performance of the models on UMAD to detect anomalies. 
For the multi-class task, we further report the mIoU and macro-F1 which represent the average IoU and F1-score across all classes, respectively.





\begin{figure*}[thbp]
\centering
\includegraphics[scale=0.0485]{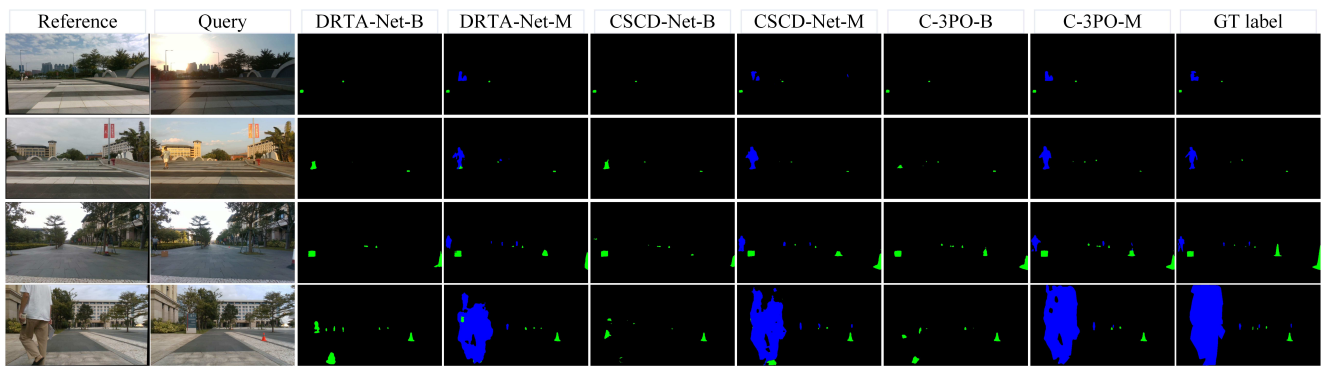}  
\caption{Qualitative comparison results of binary and multi-class anomaly detection with reference. -M and -B indicate multi-class and binary ADr, respectively. \textcolor{blue}{Blue} color represents dynamic objects, while \textcolor{green}{green} color denotes anomalous objects.}
\label{img4.1}
\end{figure*}

\subsubsection{Implementation details}
To ensure a fair comparison, we employ the same training strategies as described in~\cite{wang2023reduce} to train all the models in this experiment. It should be noted that in order to expedite the training process, we resize the images to a resolution of 360 $\times$ 640. All models in our experiments are trained on an NVIDIA GeForce RTX 3090 GPU with a memory capacity of 24GB. The results reported for the models are obtained by running the official source codes with the default settings provided. Note that the model with the best performance on the validation set is selected to evaluate the generalization performance of the model.
In the case of the multi-class anomaly detection task, we modify the default output classes from 2 to 3 to suit our experimental requirements.

We benchmarked the following representative change detection models: FC-Siam~\cite{daudt2018fully}, ChangeNet~\cite{varghese2018changenet}, CSCDNet~\cite{sakurada2020weakly}, DR-TANet~\cite{chen2021dr}, C-3PO~\cite{wang2023reduce} in both binary and multi-class ADr tasks. 
For the binary ADr, change detection models can directly detect anomalous changes while ignoring common changes, such as dynamic objects. For the multi-class ADr, change detection models detect all the changes and distinguish common and anomalous changes. As shown in Table \ref{tab:4.2}, for most of the compared methods, there is a significant performance improvement from binary ADr to multi-class ADr. This is because, as shown in Fig. ~\ref{img4.1}, the model can further distinguish common changes (e.g., individuals in the 2nd and 4th row) after giving the pattern of the common changes in training, resulting in more accurate anomaly predictions.

\section{Conclusions}
\label{sec:5}
In this paper, we propose UMAD, the first benchmark dataset for the ADr in the context of patrol robot scenes. To address the challenge of aligning the reference and query pairs, we propose an adaptive warping method, which allows for accurate comparison between the image pairs. Unlike ADwr approaches that face difficulties in handling the diverse range of anomalies, ADr offers an alternative by leveraging the change detection between the reference and query image pairs. By utilizing the UMAD dataset, we have conducted evaluations of several state-of-the-art scene change detection methods to identify anomalies and establish a baseline performance. We intend that UMAD will facilitate the development of more effective ADr methods, thereby bridging the existing gap in this research domain.










\bibliographystyle{IEEEtran}
\bibliography{root}

\end{document}